# Prompt Learning for Oriented Power Transmission Tower Detection in High-Resolution SAR Images

Tianyang Li [a,b,c], Chao Wang [a,b,c,*], Fan Wu [a,b], Yixian Tang [a,b,c], Lu Xu [a,b], Hong Zhang [a,b,c]

[a] Key Laboratory of Digital Earth Science, Aerospace Information Research Institute, Chinese Academy of Sciences, Beijing 100094, China;

[b] International Research Center of Big Data for Sustainable Development Goals, Beijing 100094, China;

[c] College of Resources and Environment, University of Chinese Academy of Sciences, Beijing 100049, China;

* Correspondence: wangchao@radi.ac.cn (C.W.);

## Abstract

Detecting transmission towers from synthetic aperture radar (SAR) images remains a challenging task due to the comparatively small size and side-looking geometry, with background clutter interference frequently hindering tower identification. A large number of interfering signals superimposes the return signal from the tower. We found that localizing or prompting positions of power transmission towers is beneficial to address this obstacle. Based on this revelation, this paper introduces prompt learning into the oriented object detector (P2Det) for multimodal information learning. P2Det contains the sparse prompt coding and cross-attention between the multimodal data. Specifically, the sparse prompt encoder (SPE) is proposed to represent point locations, converting prompts into sparse embeddings. The image embeddings are generated through the Transformer layers. Then a two-way fusion module (TWFM) is proposed to calculate the cross-attention of the two different embeddings. The interaction of image-level and prompt-level features is utilized to address the clutter interference. A shape-adaptive refinement module (SARM) is proposed to reduce the effect of aspect ratio. Extensive experiments demonstrated the effectiveness of the proposed model on high-resolution SAR images. P2Det provides a novel insight for multimodal object detection due to its competitive performance.

## Keywords

Synthetic aperture radar (SAR), Power transmission tower, Oriented object detection, prompt learning, Multimodal learning.

## 1. Introduction

Power transmission towers constitute crucial and extensively dispersed infrastructure within the power industry, rendering them highly vulnerable to extreme weather conditions. As components of power transmission, these towers are situated in suburban, farmland, forest, and other diverse natural environments (Li et al. 2022a). In such scenarios, transmission towers are exposed to various natural disaster risks, including flooding, landslides, windstorms, etc. (Liu et al. 2019; Zhang and Xie 2019; Zhou et al. 2022b). Efficient detection of transmission towers is crucial for ensuring safe operation and safety monitoring. Electric power companies and government departments must ensure uninterrupted power supply and the faultless operational status of transmission towers. Remote sensing serves as an effective means for monitoring power transmission towers (Matikainen et al. 2016). Over an extended period, aerial remote sensing, primarily conducted by unmanned aerial vehicles (UAVs), has been the principal method for this task due to its exceptional maneuverability and high resolution (Bian et al. 2019; Deng et al. 2014). Conducting comprehensive inspections of transmission towers on a large scale using UAVs is challenging because of the constraints imposed by battery technology (Galkin et al. 2019). Hence, achieving rapid and accurate detection of power transmission towers on a large scale constitutes a significant research challenge.

Surveys of power transmission towers must encompass extensive areas, and satellite remote sensing technology presents appealing alternatives. Numerous studies have been published on the application of satellite remote sensing methods in detecting power transmission towers (Haroun et al. 2020; Li et al. 2022a; Zha et al. 2023). Images captured by visible and near-infrared sensors exclusively depict the energy reflected from the Earth's surface and are unattainable in darkness. These images are vulnerable to adverse weather conditions, such as clouds and rain, potentially hindering their utility. Synthetic Aperture Radar (SAR) can capture images under diverse weather conditions, making it suitable for disaster monitoring applications compared to passive sensors. It records backscatter signals from the target through the emission of microwave radiation. From a satellite perspective, a power transmission tower is a compact and vertically structured object that typically remains unchanged over time. The side-looking geometry of SAR images aids in detecting these vertical features (Yan et al. 2011). However, small features are frequently associated with geometric deformations and multi-path scattering. The inherent speckle noise in SAR images induces pseudo-random fluctuations in radar intensity. Additionally, the geometric distortion of the image introduces interference from background clutter. Researchers are increasingly focusing on the detection of power transmission towers from SAR images as the image resolution improves.

Currently, numerous studies have been conducted on the detection of power transmission towers from SAR images. The majority of methods rely on intensity information, geometric features, and interference information. In the early stages, constant False Alarm Rate (CFAR) and Support Vector Machine (SVM) were commonly employed to analyze intensity information from power transmission towers (He et al. 2013; Liu et al. 2012; Zhang et al. 2013b). Alternative methods leverage geometric features to identify power transmission towers (Zhang et al. 2013a). The proposal of the polar coordinate semi-variogram aims to extract geometric features from regions of interest and decrease false alarms (Zeng et al. 2017). Distribution patterns of pixels have been devised to detect transmission towers, including Signal-to-Clutter Ratios (SCRs) (Li et al. 2022a). Additionally, multi-baseline SAR interferometry-correlated synthesis images are employed to mitigate background clutter caused by mountainous layover (Wu et al. 2023). Traditional methods face challenges in addressing power transmission tower detection across diverse scenarios. Presently, deep learning has significantly advanced in the field of power transmission tower detection. Classical object detection networks, including YOLO and SSD, have proven successful in extracting power transmission towers from high-resolution SAR images (Gao et al. 2019; Tian et al. 2020). Subsequently, oriented bounding boxes are implemented to mitigate the interference from background clutter (Li et al. 2023). However, achieving high-precision power transmission tower detection in SAR images remains a challenging task. The majority of existing detectors overlook the multimodal approach in dealing with disturbances in complex backgrounds.

Previous research has made substantial contributions to fully exploiting the features of power transmission towers in SAR images. However, in practical scenarios, power transmission towers are extensively distributed, and diverse terrains substantially impact detection efficiency. As shown in Fig. 1, varying orientations of power transmission towers lead to signal overlap with other distributed scatters. In addition to the side-looking geometric features, the positions of power transmission towers typically remain unchanged over time. A significant amount of prior knowledge about the locations is already available, including those marked in the OpenStreetMap (OSM). The current research has not fully harnessed multimodal data to optimize the detection task in SAR images.

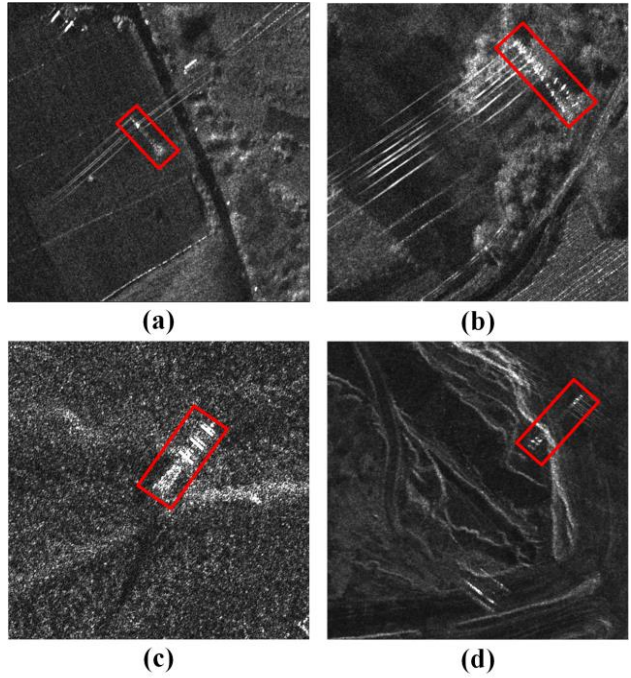

Fig. 1. Power transmission tower at farmland (a), woodland (b), deserts (c), mountains (d) on SAR imagery. The red boxes mark the exact location of each tower. Background features cause different levels of interference.

To address the issues mentioned earlier, this paper introduces a novel prompt-based detector, P2Det, designed for detecting oriented power transmission towers in SAR images. P2Det is a multimodal approach that integrates image and positional prompts. Specifically, a sparse prompt encoder (SPE) is developed to encode point-wise prompts using Fourier feature mapping. Subsequently, a two-way fusion module (TWFM) is introduced to learn the interrelationship between the data of the two modalities. This approach effectively addresses the challenge of heterogeneous data fusion which is called multimodal data fusion module (MDF). Moreover, a shape-adaptive refinement module is developed to comprehensively consider the influence of shape information in sample selection.

The main contributions of this paper can be summarized as the following:

(1) A novel prompt learning-based detector P2Det is designed for power transmission tower detection on SAR images through the fusion of point prompts and a shape-adaptive sample selection strategy.

(2) Demonstrating the effectiveness of prompt learning in object detection at fixed positions, we propose a sparse prompt encoder and a two-way fusion module to utilize point prompts for guiding the classification.

(3) We innovatively propose a shape-adaptive refinement strategy that dynamically selects samples

based on shape information and feature distribution, thereby further improving performance.

## 2. Related work

In this section, we summarize the works of oriented object detection and prompt learning, including key models and trends. Then the Transformer-based methods and their development are described briefly.

### 2.1 Oriented object detection

Oriented object detection has been a crucial focus within the field of remote sensing. Over the past decades, this field has attracted significant attention owing to its broad range of applications (Ding et al. 2019; Han et al. 2021; Li et al. 2022b; Xie et al. 2021b; Yang et al. 2019; Zhou et al. 2022c). Unlike the horizontal bounding box (HBB) (Redmon et al. 2016; Ren et al. 2015), the oriented bounding box (OBB) can represent objects in any direction. The objects in remote sensing images are observed from an overhead perspective with arbitrary orientations. Currently, there are several noteworthy studies on oriented object detection. Some efforts aim to address challenges associated with angles, scales, and aspect ratios by establishing dense anchor boxes to achieve enhanced results (Ma et al. 2018; Yang et al. 2019). RoI Transformer designs a rotated region of interest (RoI) to mitigate the imbalance of the preset horizontal anchors (Ding et al. 2019). On the other hand, $S^2A$-Net and $R^3$Det implement feature alignment to improve classification and regression tasks through refinement heads (Han et al. 2021; Yang et al. 2021a). These works optimize the localization offsets of the bounding boxes during the feature refinement stage.

Some works also focus on regression loss for oriented object detection. The angle representations in OBB are periodic, leading to loss oscillations (Yang and Yan 2020) and problems of representation ambiguity (Ming et al. 2021). Circular smooth label transforms orientation regression into a classification task to address the out-of-bounds angle effect (Yang and Yan 2020). Subsequently, Yang et al. proposed alternative approaches to measure the distance between OBBs, thereby avoiding the issue of angular regression (Yang et al. 2021b; Yang et al. 2021c). They employed Gaussian Wasserstein distance and Kullback-Leibler divergence to transform the representation of the bounding box into a two-dimensional Gaussian distribution, thereby resolving the inconsistency between the loss and the metric, as well as addressing the regression boundary problem.

### 2.2 Prompt learning

The inception of prompt representation learning can be attributed to the vision-language pretrained model (Zhou et al. 2022a). CLIP illustrates the viability of training transferable visual models from raw text describing images (Zhou et al. 2022a). SAM, as a promptable model, is capable of zero-shot transfer to new image distributions (Kirillov et al. 2023). Points, bounding boxes, masks, and text serve as input patterns for prompting learning tasks. SAM's performance can be comparable to or even surpass that of previous fully supervised fine-tuning models. Owing to its outstanding performance in computer vision downstream tasks, the promptable architecture introduces a new direction in the field of artificial intelligence. Multi-modal prompt learning is suggested to enhance the alignment between vision and language representations (Khattak et al. 2023).

Several notable works focus on prompt-based object detection. Researchers are dedicated to adapting pre-trained visual language models for addressing downstream recognition tasks (He et al. 2023; Long et al. 2023; Lu et al. 2022; Nie et al. 2023). The proposed prompt representation learning in the image classification task proves suboptimal when applied to the object detection task. Even a slight modification in the prompt can ultimately result in a substantial positive or negative impact on detection outcomes. To address this issue, DetPro learns continuous prompt representations for open-vocabulary object detection (Du et al. 2022). Furthermore, assigning suitable strategies for various modal variable inputs is challenging; therefore, UniSOD introduced modality-aware prompts for adaptive prompt learning (Wang et al. 2023).

## 2.3 Vision Transformer

The Transformer exhibited outstanding performance in natural language processing by capturing long-range dependencies among sequence elements. Subsequently, it was applied to image classification by converting image patches into sequences, marking the first successful implementation in computer vision (Dosovitskiy et al. 2020). Leveraging its exceptional global interaction capability, the Vision Transformer attains state-of-the-art performance in various computer vision tasks. Enhancements to the Vision Transformer primarily center around the following aspects: training strategies (Chen et al. 2021a; Touvron et al. 2021), computation complexity (Liu et al. 2021; Xie et al. 2021a), and model architecture (Carion et al. 2020; Wang et al. 2021, 2022; Zhang et al. 2022; Zhu et al. 2020). DETR pioneered end-to-end object detection, eliminating the need for manually designed modules in traditional detectors (Carion et al. 2020). Adaptations to dense prediction tasks were proposed, such as using a shifted window strategy to interact with cross-window information (Liu et al. 2021), and incorporating a pyramid structure for extracting multi-scale features (Wang et al. 2021, 2022). Furthermore, the hybrid architecture of CNN and Transformer exhibits certain potential by explicitly modeling local and global contexts (Chen et al. 2021b).

## 3. Methodology

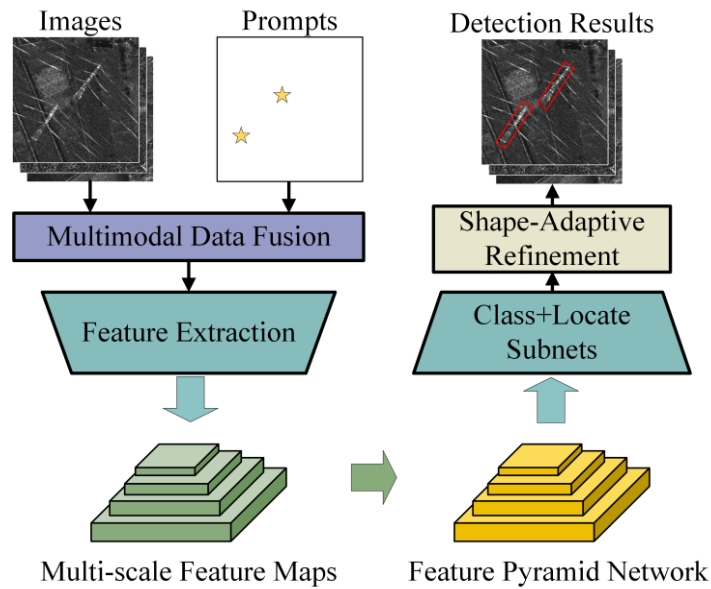

Fig. 2. The overall structure of the proposed P2Det.

Considering the side-looking geometry and the time-invariant positional properties, we proposed an object detection framework for power transmission towers in SAR images that integrates positional prompts. In contrast to detection algorithms for mobile targets, we innovatively introduced point-wise prompts labeled from the OSM to build a novel detector, termed P2Det. Illustrated in Fig. 2, P2Det adheres to the design principles of single-stage oriented object detection. Our model comprises three components: multimodal data fusion (MDF), feature extraction, and subnetworks for classification and regression. Initially, the input SAR images and points are encoded into embeddings during the data fusion stage. Secondly, the Pyramid Vision Transformer (PVT) serves as the backbone for extracting multi-scale features. Third, the Feature Pyramid Network (FPN) decodes feature maps, with each layer connecting the classification and regression subnetworks to predict categories and locations. Lastly, a shape-adaptive refinement module is devised to generate quality measurements for positive samples using normalized shape distance. The following subsections in this section detail the proposed modules.

## 3.1 Multimodal Data Fusion Module

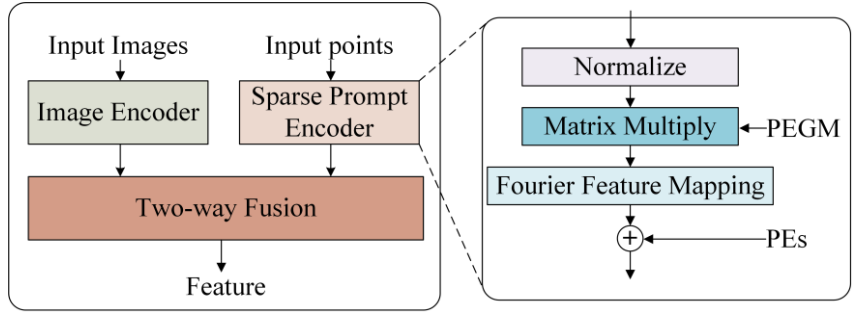

Fig. 3. The structure of the multimodal data fusion method. PEGM and PE represent Positional Encoding Gaussian Matrix and Points Embeddings respectively.

The multimodal data fusion module addresses the disparities between heterogeneous data using Transformer. Data from distinct sources, such as images and points, cannot be directly fused. To address this issue, we propose two encoders to generate embeddings for fusion. In the image encoder, a lightweight Transformer is employed to embed images. As points are discrete data, the sparse prompt encoder (SPE) represents points by learned embeddings summed with positional encodings. The image embedding and prompt embedding are subsequently mapped to a feature map by the two-way fusion module (TWFM), as shown in Fig. 3. The subsequent paragraphs provide a detailed description of MDF.

The image encoder framework aligns with the Vision Transformer (Dosovitskiy et al. 2020) and comprises four modules: patch embedding, positional embedding, Transformer encoder, and the neck layer. First, images are partitioned into sequence patches, followed by dimensional mapping and flattening into one-dimensional embeddings. To preserve positional information, positional embeddings are added to patch embeddings before they are fed into Transformer encoders. Lastly, the output channels are adjusted by multiple convolutional layers to produce the final image embeddings.

The dimensions of the point prompts are considerably smaller than those of the images, owing to the distinct data types. Thus, the points are represented by learned embeddings, which are summed with positional encodings to address the disparities in heterogeneous data. Multimodal data fusion typically transforms and aligns diverse data to enable uniform operations. Fourier features enable the network to learn high-frequency functions in low-dimensional domains (Tancik et al. 2020). Consequently, the Fourier feature mapping is applied to generate sparse embeddings. As shown in Figure2, the sparse prompt encoder initially normalizes point prompts to prevent excessive gradients. Prior to mapping, the position v is expressed in terms of spatial coordinates multiplied by the distribution of the Gaussian vector b. Subsequently, the Fourier feature mapping fuction $\gamma$ featurizes the input coordinates (Equation 1). The final sparse embeddings are composed of the mapped spatial coordinates and a set of learnable embeddings.

$$\gamma(v) = \left[a_1 \cos(2\pi b_1^T v), a_1 \sin(2\pi b_1^T v), \ldots, a_m \cos(2\pi b_m^T v), a_m \sin(2\pi b_m^T v)\right]^T \quad (1)$$

The preceding two encoders have generated image embeddings and sparse embeddings. As shown in Fig. 4, this module assembles cross-attention Transformer blocks to discern the relationship between the two embeddings. As is commonly understood, a Transformer block's input necessitates three matrices: Q, K, and V (Equation 2,3). Typically, these matrices originate from the same input features through the fully connected layer. In the realm of multi-modal fusion, conventional Transformers are no longer adaptable to multiple types of inputs. In contrast to traditional solutions, the Q, K, and V of the cross-attention Transformer are derived from different inputs (Equation 4,5). The sparse prompt embedding and the image embedding, summed with positional encoding, serve as input to the cross-attention Transformer block. Then the final output feature is up-sampled to the original resolution using bilinear interpolation.

$$\begin{cases} Q = \text{proj}(x) \\ K = \text{proj}(x) \\ V = \text{proj}(x) \end{cases} \quad (2)$$

$$\text{Attn}(Q, K, V) = \text{Softmax}\left(\frac{QK^T}{\sqrt{d_{\text{head}}}}\right)V \quad (3)$$

$$\text{Attn}_{t2i} = \text{Attn}(q, k, k_{pe}) \quad (4)$$

$$\text{Attn}_{i2t} = \text{Attn}(k, q, q_{pe}) \quad (5)$$

Where, x refers to the input feature. The $\text{proj}(\cdot)$ is a fully connected layer, $\text{Attn}_{t2i}$ and $\text{Attn}_{i2t}$ represent the cross-attention of token and image to each other, respectively. Variables $q$, $k$, $q_{pe}$, and $k_{pe}$ denote the token, image embeddings and the sum of the first two with the corresponding positional embeddings.

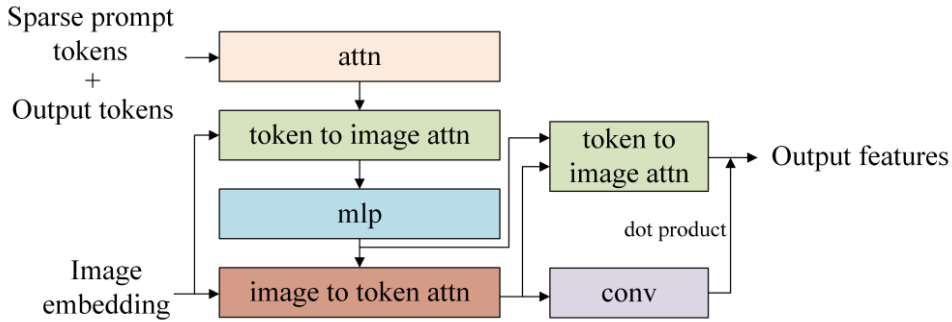

Fig. 4. The structure of the TWFM. Modules attn, conv, and mlp represent the attention layer, the convolutional layer, and the multilayer perceptron, respectively. TWFM adopts cross-attention between tokens and images.

## 3.2 Shape-adaptive refinement module

The shape of a power transmission tower in the image is intricately linked to the angle of incidence, particularly the aspect ratio. A low incidence angle results in a larger aspect ratio, indicating variations in shapes from near to far range within the same image. This discrepancy becomes more pronounced in other SAR imaging modes and image acquisitions. Certain prior studies attempted to utilize aspect ratio information for filtering geometric parameters (Li et al. 2022a), but still faced the limitations of manual design. To mitigate shape effects, we introduced a dynamic sample selection strategy into P2Det, termed the shape-adaptive refinement module (SARM). SARM proves to be an effective and flexible label assignment method (Fig. 5). Sample selection involves two aspects: shape sensitivity thresholds and positive sample quality assessment.

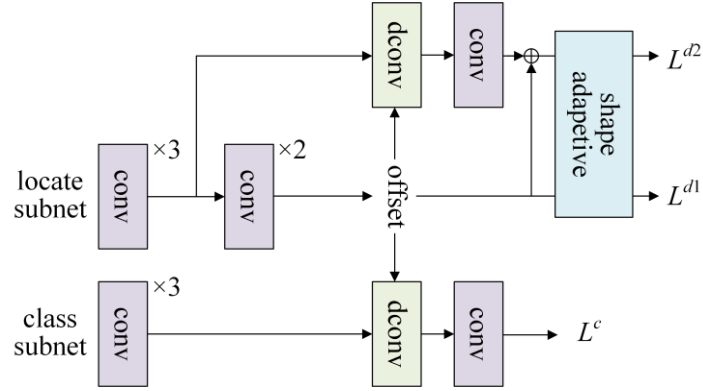

Fig. 5. The structure of the detection head. Module dconv refers to the deformable convolutional layer. $L^c$, $L^{d1}$, and $L^{d2}$ are loss functions for classification and localization.

Some studies have investigated the correlation between aspect ratio and threshold setting in label assignment (Hou et al. 2022; Zhang et al. 2020). A lower intersection of union (IoU) threshold performs better when the aspect ratio of the object is large. Hence, the selection strategy for positive and negative samples can be enhanced by incorporating aspect ratio information. Thus, the dynamic effect is that the larger the aspect ratio, the lower the IoU threshold of the object. As demonstrated in Equation 6, the mean and standard deviation of objects are calculated to dynamically adjust the threshold.

$$\mu = \frac{1}{n}\sum_{j=1}^{n} IoU_{i,j}, \qquad \sigma = \sqrt{\frac{1}{n}\sum_{j=1}^{n} (IoU_{i,j} - \mu)^2} \qquad (6)$$

Where, n is the number of candidate samples, and $IoU_{i,j}$ is IoU threshold between the i-th ground truth box and the j-th prediction.

We define the $\alpha$ as the ratio of the long edge to the short edge. As mentioned before, the IoU threshold decreases with increasing aspect ratio. Therefore, the adaptive function $f(\cdot)$ is designed as a monotonically decreasing function as the follow equation. The shape factor $w$ is added to the aspect ratio as an adjustable parameter.

$$f(\alpha_i) = e^{-w\alpha_i} \qquad (7)$$

For the i-th ground-truth box, the IoU threshold $T_i$ is calculated as:

$$T_i = f(\alpha_i) * (\mu + \sigma) \qquad (8)$$

For SAR images, background clutter significantly affects the detection accuracy of power transmission towers. Pixels within the towers, especially those near strong scattering areas, are more representative of the tower's features than those near the boundaries or in weaker scattering areas. Therefore, to mitigate background clutter interference, we propose the use of normalized shape distance in the quality assessment stage for positive samples. The quality of each sample point correlates with the shape and the scattering center. The distance from a sample to the corresponding object's center is determined using shape information. This position, relative to the object, is estimated to facilitate the calculation of the normalized shape distance $D_{ij}$ (equation 9).

$$D_{ij} = \sqrt{\frac{(x_i - x_j)^2}{w_i} + \frac{(y_i - y_j)^2}{h_i}} \tag{9}$$

Where, $x$, $y$, $w$, and $h$ are the center coordinates, width, and height of the ground-truth box. The subscripts $i$ and $j$ represent the i-th sample point and the j-th object center, respectively.

To reduce the weight of the samples at the boundary, we use the same designed function as $f(\cdot)$ to describe the sample quality $Q_{ij}$:

$$Q_{ij} = e^{-D_{ij}} \tag{10}$$

## 3.3 Loss function

The total loss $L_{total}$ comprises three parts: the classification loss $L^c$, the initial detection head loss $L^{d1}$, and the refinement detection head loss $L^{d2}$. By assigning corresponding weight coefficients $\lambda_1$, $\lambda_2$, and $\lambda_3$ to each term, $L_{total}$ is calculated as:

$$L_{total} = \lambda_1 L^c + \lambda_2 L^{d1} + \lambda_3 L^{d2} \tag{11}$$

The classification loss $L^c$ uses the focal loss (Lin et al. 2017). For the i-th object, the regression loss $L^{d1}$ is calculated as:

$$L_i^{d1} = \frac{1}{N} \frac{1}{\sum_{p_j \in P} Q_{ij}} \sum_{ij} Q_{ij} l_{ij}^{reg} + l_{ij}^{bc} \tag{12}$$

Where, $N$, $j$, and $P$ refer to the total number, the index, and the set of the predicted convex hulls, respectively. $l_{ij}^{reg}$ adopts the GIoU loss (Union 2019). Boundary-center loss (Hou et al. 2022) $l^{bc}$ is employed to constrain the boundary and center points.

In the refinement detection head, the regression loss $L_i^{d2}$ for each convex hull is defined as:

$$L_i^{d2} = \frac{1}{N} \frac{1}{\sum_{p_j \in P} Q'_{ij}} \sum_{ij} Q'_{ij} l_{ij}^{reg} \tag{13}$$

In the detection head loss function, $Q_{ij}$ and $Q'_{ij}$ are the aforementioned quality coefficients for each positive sample. And the adaptive weight is assigned to each sample based on $Q$.

# 4. Experiments

## 4.1 Datasets

Extensive experiments were conducted on a dataset comprised of Capella SAR images. Owing to the absence of a publicly available and verifiable dataset for power transmission towers, we manually constructed a dataset featuring multiple angles and diverse backgrounds using high-resolution SAR data. Utilizing the Capella Space Open Data Gallery, we collected single-polarized (HH), X-band SAR data across various imaging modes. The images were annotated leveraging expert knowledge and Open Street Map (OSM) data. Bounding boxes for power transmission towers were delineated using four corner points. To incorporate location prompts into the SAR images, we extracted and adapted tower-related fields from the OSM for power transmission tower objects. Most tower locations in the OSM are typically marked at the base. Consequently, point prompts were accurately positioned as near as possible to the base of the power transmission towers to mitigate the effects of the SAR system's side-looking imaging mechanism.

The collected high-resolution SAR images were subsequently cropped into 512×512 pixel patches following linear stretching. The dataset comprises a total of 1614 images, featuring 2066 instances of power transmission towers. Scenarios within the dataset encompass mountains, plains, grasslands, farmlands, and urban areas. These diverse scenarios enhance the generalizability of transmission tower identification. Furthermore, the dataset's power transmission towers contain scattering characteristics across a range of incidence angles, from 25 to 50 degrees.

## 4.2 Experimental setting

The default input image size for training and validation in our experiments is set at 512×512. Based on the dataset's statistical metrics, the mean and variance for image normalization are 45.52 and 28.36, respectively. Image augmentation involves flip operations in three directions: horizontal, vertical, and diagonal, each assigned a probability of 25%. probability. All models were trained using the RTX 2080 Ti GPU, with the batch size set at 4. Training iterations were set at 120 epochs for each experiment. The SGD optimizer was employed, with an initial learning rate of 0.0025 set by default.

Average precision (AP) and average recall (AR) were employed to evaluate the performance of all models. $AP_{50}$ denotes average precision at an IoU threshold of 0.5. $AP_{75}$ represents average precision at an IoU threshold of 0.75. Given the importance of recall for power transmission tower detection, it is considered as one of the evaluation metrics. Similarly, the subscript indicates the IoU threshold used for computing the recall.

# 5. Experimental results

## 5.1 Ablation study

**Component-wise ablations**

Ablation experiments are conducted to validate the effectiveness of the proposed modules. Given that the TWFM is predicated on the SPE, the control group utilized PVT as the baseline for comparison. All modules of the P2Det framework are detailed in Table 1. The baseline model employs the RetinaNet architecture, with its backbone replaced by the PVT. As shown in Table 1, the baseline model reaches the $AP_{50}$ of 0.882, and $AR_{50}$ of 0.909.

The incorporation of MDF in prompt learning results in a 2.1% improvement in both $AP_{50}$ and $AR_{50}$. This demonstrates the efficacy of fusing prompts and images during the feature extraction phase. Subsequently, SARM contributes to an improvement of 1.9 points and 5.6 points in $AP_{50}$ and $AR_{50}$, respectively. In contrast to the baseline model, which directly classifies and regresses features, the refinement phase in SARM enhances detection accuracy. This marked increase in AP indicates the model's heightened sensitivity to shape geometry. Variations in aspect ratio, attributable to differences in incidence angles, impact the model's performance. SARM mitigates this sensitivity by employing a dynamic threshold to select high-quality samples. Ultimately, the combined application of MDF and SARM further elevates performance, resulting in improvements of 2.4% in $AP_{50}$ and 7.8% in $AR_{50}$.

**Different backbones**

Tables 2 and 3 illustrate the detection accuracy based on different backbones and heads. ResNet and PVT are recognized as classic and efficient backbones for object detection, representing the convolutional and transformer architectures, respectively. We conduct comparisons among various backbone structures to demonstrate the improvement of prompt learning in terms of feature extraction. Interestingly, both PVT and its improved version in Table 2 exhibit a slight reduction in detection accuracy. This phenomenon could be attributed to the fact that the Transformer requires more training samples. MDF addresses this limitation by utilizing location information. In comparison to PVT2, the metrics of $AP_{50}$ and $AR_{50}$ are enhanced by 2.2% and 2.1%, respectively, in our proposed method.

The MDF module plays a pivotal role in the implementation of prompt learning in our methods. Prompt learning significantly enhances the power transmission tower characterization. Our module improves performance on feature localization through the fusion of multimodal data. The output from the backbone critically determines the performance of subsequent detection and classification sub-networks. The power transmission tower, comprising pylons, power lines, insulators, and more, is depicted on the high-resolution SAR image at the appropriate angle of incidence. To elucidate the feature extraction process of these components, we visualize the feature map with and without the MDF module using explainable deep learning techniques (Muhammad and Yeasin 2020). As shown in Fig. 6, MDF significantly improves the representation of the characteristics of transmission towers. The MDF is adept at recognizing subtle features such as conductors and insulators while focusing on the pylons of the tower. The high interpretability of these components facilitates the subsequent detection of the network. In complex scenarios, MDF efficiently extracts pylon features, even in the presence of strong scatters.

In summary, our MDF is designed for precise localization and proves effective in identifying the characteristics of various power transmission tower assemblies.

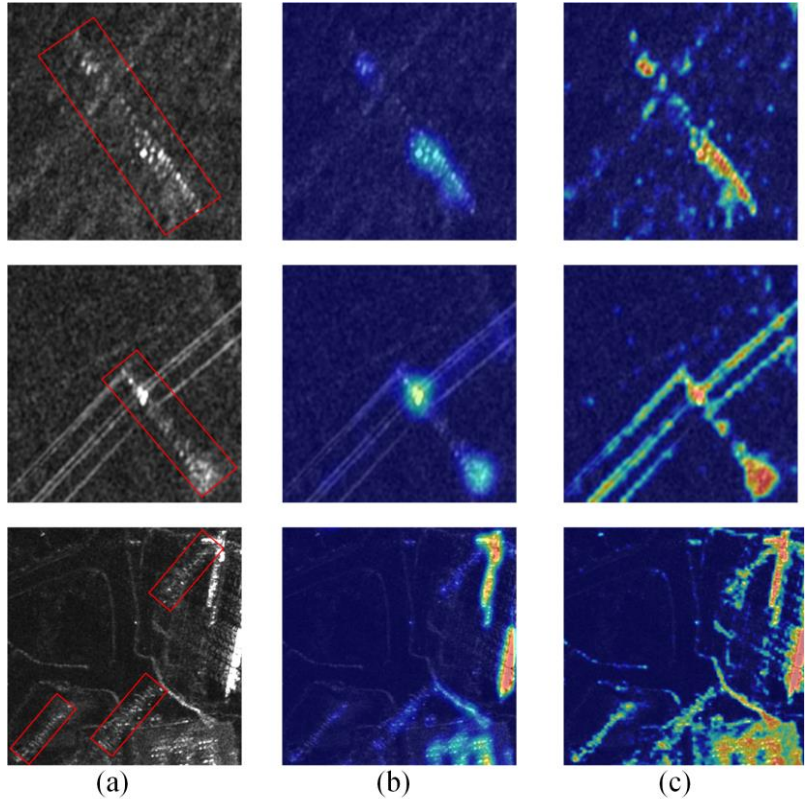

Fig. 6. The visualization of features among the baseline (b) and our model (c). Compared to the baseline model, our method presents more discriminative representations, including the pylon, conductors, and insulators.

**Different detection heads**

We evaluate the impact of anchor-based and anchor-free methods on model performance. The baseline model incorporates the detection head from RetinaNet. The Feature Refinement Module (FRM) is derived from the detection head of the R3Det, introducing a feature refinement operation inspired by the architecture of RetinaNet. The reduced accuracy of FRM compared to the baseline suggests the limited efficacy of multiple IoU thresholds in the sample selection strategy during prediction box refinement. The IoU threshold, a crucial hyperparameter for objects of varying scales, requires careful tuning as values too high or too low may result in omission errors. The implementation of Adaptive Training Sample Selection (ATSS) underscores the significance of the threshold in sample selection, enhancing FRM by 1.9% and 1.3% for $AP_{50}$ and $AR_{50}$, respectively. While ATSS accounts for the mean and standard deviation of the ground-truth boxes, it overlooks the individual quality of each sample. Our SARM employs normalized shape distance to evaluate sample quality. SARM achieves an improvement of 2.4 points and 1.1 points on $AP_{50}$ relative to the anchor-based (FRM) and anchor-free (RepPoints) methods, respectively. Our module excels in reducing omission errors, attaining a recall of 0.943, which is higher than that achieved by other methods.

We additionally performed experiments to determine the optimal hyperparameters for the shape factor (refer to Equation 7). By analyzing statistical data to centralize the shape characteristics of power transmission towers (Fig. 7), we observed that the extreme differences in aspect ratios are within a factor of 5. Consequently, we established the shape factor w within the range (0, 4). The optimal performance was achieved when $w = 2$. The sensitivity experiments indicate that the hyperparameter shape factor is resilient across various conditions within a reasonable range.

Table 1. Evaluation of different modules in P2Det.

| PVT | MDF | SARM | $AP_{50}$ | $AR_{50}$ |
|---|---|---|---|---|
| ✓ | | | 0.882 | 0.909 |
| ✓ | ✓ | | 0.903 | 0.93 |
| ✓ | | ✓ | 0.901 | 0.965 |
| ✓ | ✓ | ✓ | **0.906** | **0.987** |

Table 2. Comparison with other feature extraction methods.

| Methods | $AP_{50}$ | $AR_{50}$ |
|---|---|---|
| ResNet-50 (He et al. 2016) | 0.895 | 0.939 |
| PVT-small (Wang et al. 2021) | 0.882 | 0.909 |
| PVT2-small (Wang et al. 2022) | 0.883 | 0.922 |
| MDF (our) | **0.905** | **0.943** |

Table 3. Comparison with other detection heads.

| Methods | $AP_{50}$ | $AR_{50}$ |
|---|---|---|
| base head (Lin et al. 2017) | **0.907** | 0.926 |
| FRM (Yang et al. 2021a) | 0.882 | 0.922 |
| ATSS (Zhang et al. 2020) | 0.901 | 0.935 |
| RepPoints (Li et al. 2022b) | 0.895 | 0.957 |
| SARM (our) | 0.906 | **0.987** |

Table 4. Analysis of different hyperparameters of the shape factor.

| Shape Factor | $AP_{50}$ | $AR_{50}$ |
|---|---|---|
| 0.25 | 0.771 | 0.909 |
| 0.5 | 0.784 | 0.926 |
| 1 | 0.792 | 0.909 |
| 2 | 0.813 | 0.935 |
| 4 | 0.791 | 0.935 |

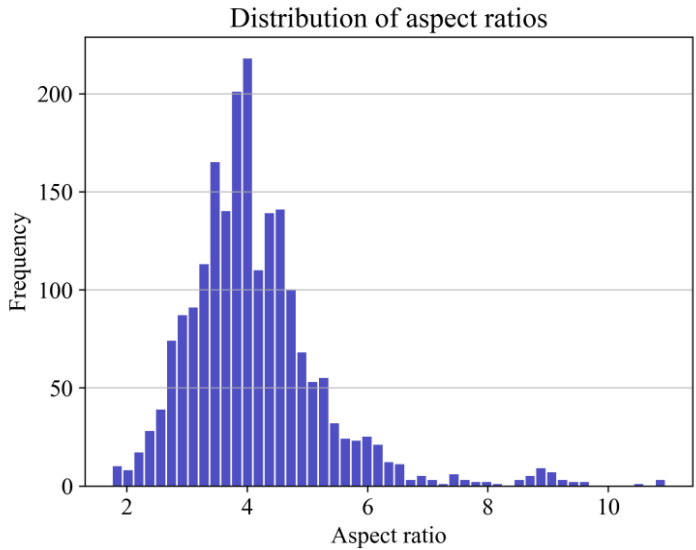

Fig. 7. Aspect ratio distribution of ground-truth boxes in the dataset.

## 5.2 Comparison with state-of-the-art models

We evaluate the performance of the current state-of-the-art (SOTA) models on the test set. As illustrated in Table 5, P2Det achieves an $AP_{75}$ of 90.3%, surpassing the performance of the compared methods. Notably, with an IoU threshold of 0.5, our recall rate reaches 99.1%, indicating a minimal rate of omissions. MDF contributes to the localization and feature extraction of power transmission towers, while SARM facilitates reliable sample selection using shape information. The results across various terrains are depicted in Figs. 8-12.

Table 5. Comparisons with different models on the dataset.

| Models | $AP_{50}$ | $AP_{75}$ | $AR_{50}$ | $AR_{75}$ |
| --- | --- | --- | --- | --- |
| R3Det (Yang et al. 2021a) | 0.897 | 0.694 | 0.95 | 0.761 |
| RetinaNet (Lin et al. 2017) | 0.904 | 0.888 | 0.968 | 0.901 |
| Oriented RepPoints (Li et al. 2022b) | 0.898 | 0.771 | 0.982 | 0.815 |
| S2A-Net (Han et al. 2021) | **0.908** | 0.882 | 0.973 | 0.901 |
| SASM (Hou et al. 2022) | 0.904 | 0.889 | 0.977 | 0.91 |
| P2Det (our) | 0.906 | **0.903** | **0.991** | **0.932** |

**Results in different regions**

Buildings are often identified as strong scattering targets in high-resolution SAR images, leading to potential confusion with scatter points from power transmission towers. As shown in Fig. 8, we present a visualization of detection results in built-up areas. The presence of high backscattering coefficients in buildings poses a significant challenge in the identification of power transmission towers. Our model achieves accurate discrimination and effectively avoids interference from complex backgrounds.

A significant proportion of power transmission towers are located in farmland areas. As shown in Fig. 9, in contrast to urban areas, farmlands in SAR images frequently exhibit organized features due to human activity. The dispersed structure of the transmission towers often leads to the challenge of distinguishing strong scattering points from field ridges or crop boundaries. An optimal incidence angle is crucial in enhancing the features of power transmission towers, thereby rendering the cables visible, as shown in Fig. 9 (c) and (e).

Additionally, interference from background clutter presents itself in various forms. Speckle noise in flat areas impairs the localization of scattering centers, as shown in Fig. 10. Cluster-like features from the canopy significantly hinder the detection of small-scale power transmission towers, as illustrated in Fig. 11. The phenomenon of perspective shrinkage results in high return characteristics, leading to layover effects in mountainous areas. Power transmission towers situated near mountains are prone to concealment by layovers, as depicted in Fig. 12.

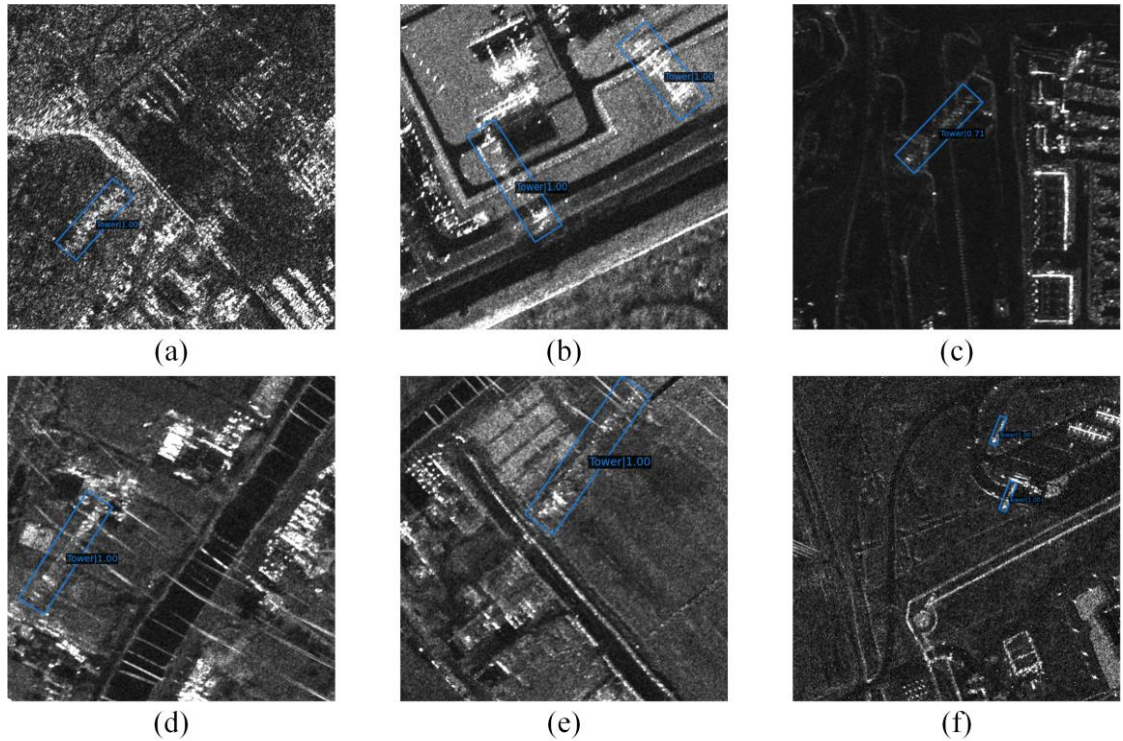

Fig. 8. Detection results in built-up areas. The blue boxes mark the locations of the power transmission towers.

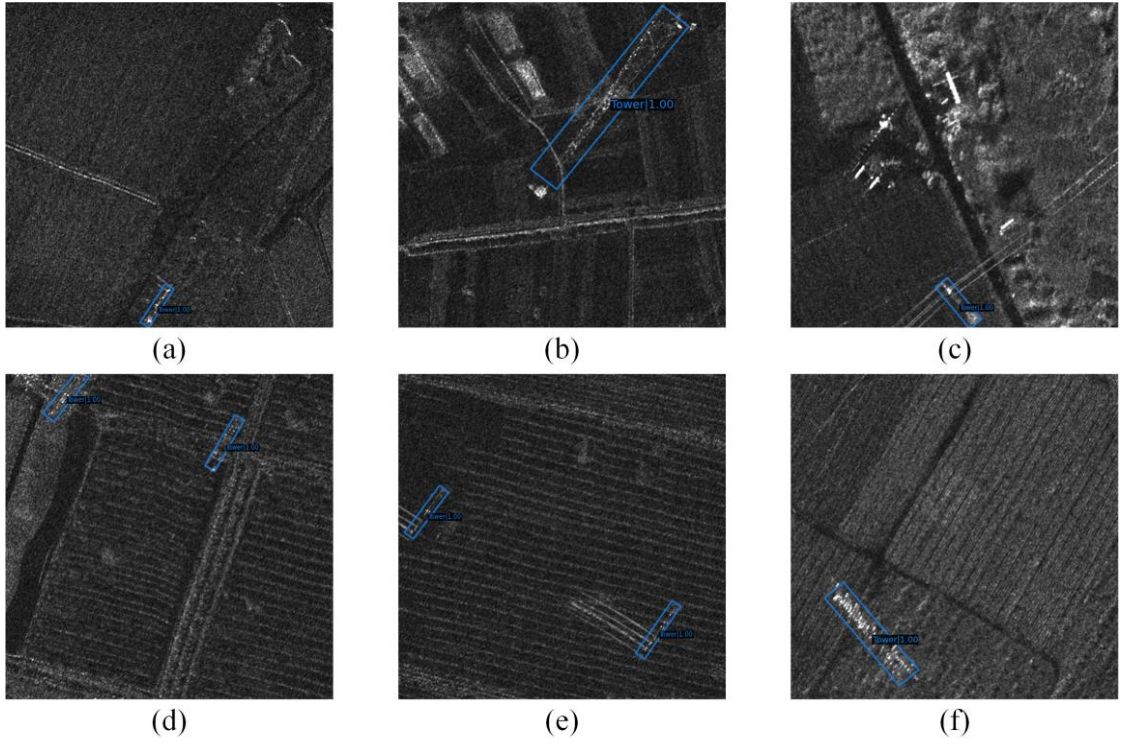

Fig. 9. Detection results in farmland areas. The blue boxes mark the locations of the power transmission towers.

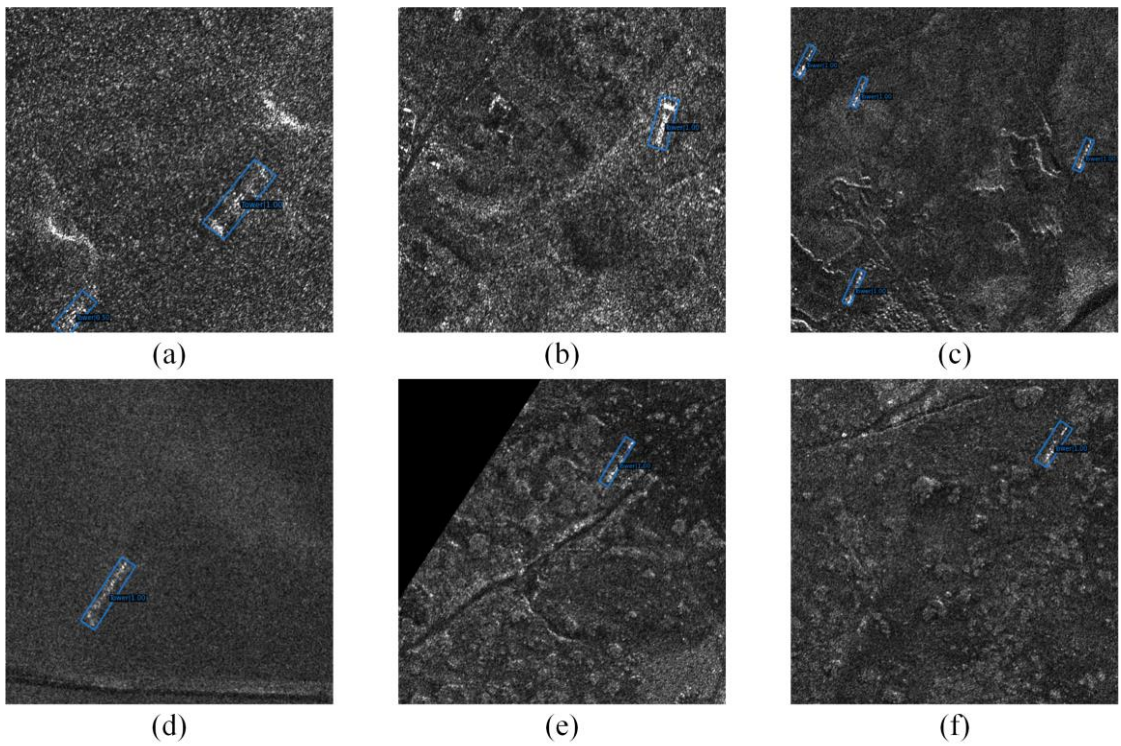

Fig. 10. Detection results in plain areas. The blue boxes mark the locations of the power transmission towers.

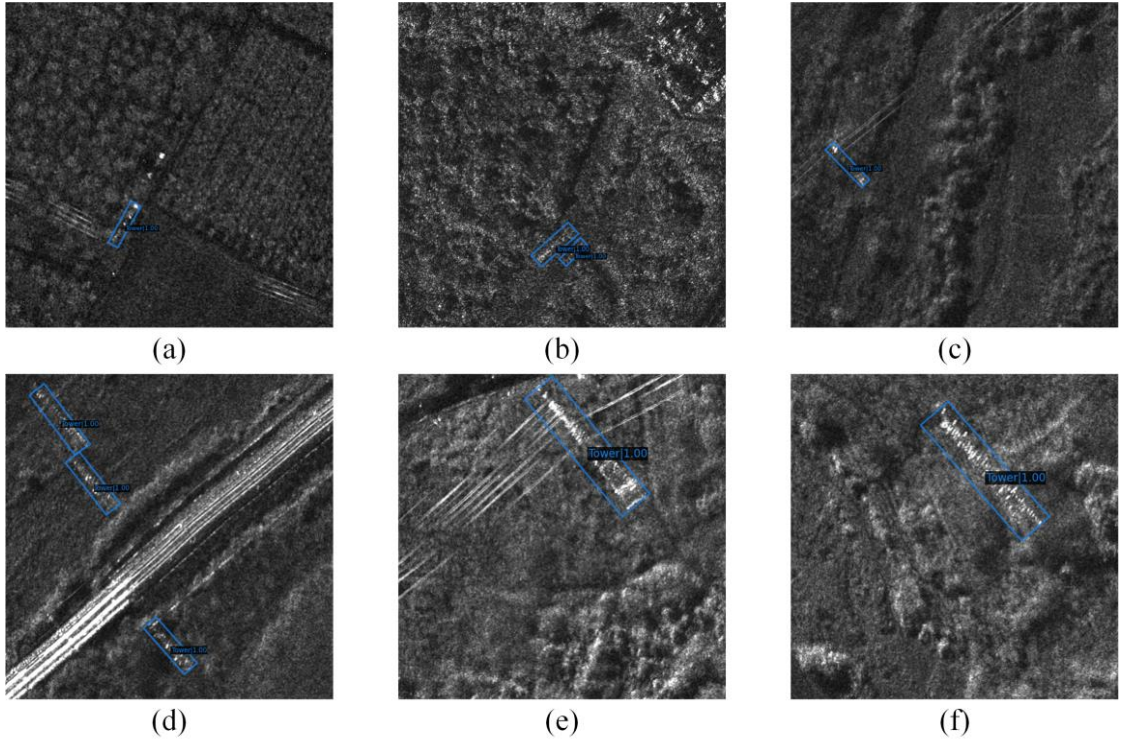

Fig. 11. Detection results in forestry areas. The blue boxes mark the locations of the power transmission towers.

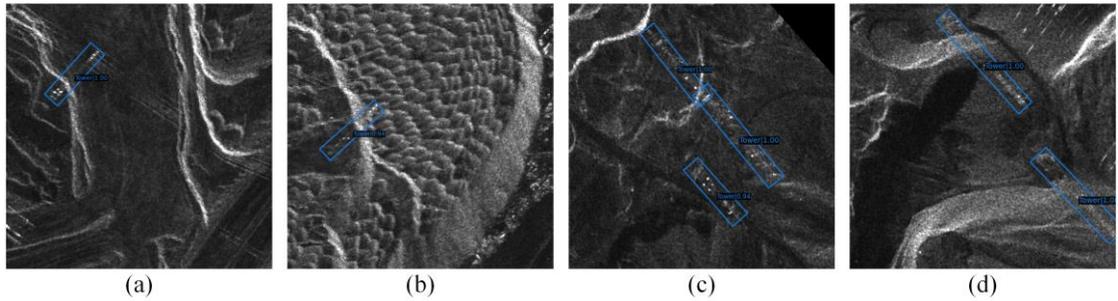

Fig. 12. Detection results in mountainous areas. The blue boxes mark the locations of the power transmission towers.

## 5.3 Discussion

Our P2Det model demonstrates optimal performance in extensive experiments. Utilizing the prompt learning technique and multimodal data, we have developed the SPE, MDF, and SARM modules to achieve performance enhancement. Specifically, Fig. 6 and Table 2 demonstrate that the MDF module in prompt learning methods significantly enhances the feature extraction capability of power transmission towers. The proposed SPE module maps point-wise positional prompts into learnable embeddings, enabling heterogeneous multimodal data fusion. MDF and SARM modules alleviate the interference from background clutter by enhancing the saliency of object features.

We introduce the shape-adaptive refinement method for sample selection. In SARM, the aspect ratio, calculated using bounding boxes, incorporates shape metrics into the IoU thresholds. Subsequently, the quality of each pixel is assessed based on the normalized shape distance. As demonstrated in Table 3, SARM greatly reduces the impact of shape variations under the condition of side-looking geometry. Furthermore, the robustness of the hyperparameters alleviates the troubles associated with parameter tuning.

# 6. Conclusion

Detecting Power Transmission towers in high-resolution SAR images is an ongoing challenge. Existing methods often struggle with background interference and typically overlook the potential of multimodal data. In this study, we introduce P2Det, a model based on prompt learning, for efficient detection of power transmission towers in SAR images. Specifically, P2Det comprises two novel modules: the Multimodal Data Fusion (MDF) module and the Shape-Adaptive Refinement Module (SARM). First, we designed a Sparse Prompt Encoder (SPE) for localization, which utilizes Fourier feature mapping. Second, the two-way fusion module (TWFM) employs cross-attention to learn the relationships within multimodal data. Third, SARM utilizes dynamic IoU thresholds and conducts quality assessments through the geometry of samples to minimize the impact of incidence angles. we design a quality factor into the loss function to enhance the sample selection strategy. P2Det demonstrates exceptional performance in high-resolution SAR image analysis. The extensive experimental results underscore the effectiveness of our model.

# Declaration of Competing Interest

The authors declare that they have no known competing financial interests or personal relationships that could have appeared to influence the work reported in this paper.

# Acknowledgements

The authors would like to thank Capella Space for providing the Capella SAR data (https://www.capellaspace.com/). This paper was funded by the National Natural Science Foundation of China, Grant No.41930110.

# Reference

Bian, J., Hui, X., Zhao, X., & Tan, M. (2019). A monocular vision–based perception approach for unmanned aerial vehicle close proximity transmission tower inspection. *International Journal of Advanced Robotic Systems, 16*, 1729881418820227

Carion, N., Massa, F., Synnaeve, G., Usunier, N., Kirillov, A., & Zagoruyko, S. (2020). End-to-end object detection with transformers. In, *European conference on computer vision* (pp. 213-229): Springer

Chen, H., Wang, Y., Guo, T., Xu, C., Deng, Y., Liu, Z., Ma, S., Xu, C., Xu, C., & Gao, W. (2021a). Pre-trained image processing transformer. In, *Proceedings of the IEEE/CVF conference on computer vision and pattern recognition* (pp. 12299-12310)

Chen, J., Lu, Y., Yu, Q., Luo, X., Adeli, E., Wang, Y., Lu, L., Yuille, A.L., & Zhou, Y. (2021b). Transunet: Transformers make strong encoders for medical image segmentation. *arXiv preprint arXiv:2102.04306*

Deng, C., Wang, S., Huang, Z., Tan, Z., & Liu, J. (2014). Unmanned Aerial Vehicles for Power Line Inspection: A Cooperative Way in Platforms and Communications. *J. Commun., 9*, 687-692

Ding, J., Xue, N., Long, Y., Xia, G.-S., & Lu, Q. (2019). Learning RoI transformer for oriented object detection in aerial images. In, *Proceedings of the IEEE/CVF Conference on Computer Vision and Pattern Recognition* (pp. 2849-2858)

Dosovitskiy, A., Beyer, L., Kolesnikov, A., Weissenborn, D., Zhai, X., Unterthiner, T., Dehghani, M., Minderer, M., Heigold, G., & Gelly, S. (2020). An image is worth 16x16 words: Transformers for image recognition at scale. *arXiv preprint arXiv:2010.11929*

Du, Y., Wei, F., Zhang, Z., Shi, M., Gao, Y., & Li, G. (2022). Learning to prompt for open-vocabulary object detection with vision-language model. In, *Proceedings of the IEEE/CVF Conference on Computer Vision and Pattern Recognition* (pp. 14084-14093)

Galkin, B., Kibilda, J., & DaSilva, L.A. (2019). UAVs as mobile infrastructure: Addressing battery lifetime. *IEEE Communications Magazine, 57*, 132-137

Gao, Y., Yang, W., Li, C., & Zou, F. (2019). Improved SSD‐based transmission tower detection in SAR images. *The Journal of Engineering, 2019*, 7161-7164

Han, J., Ding, J., Li, J., & Xia, G.-S. (2021). Align deep features for oriented object detection. *IEEE Transactions on Geoscience and Remote Sensing, 60*, 1-11

Haroun, F.M.E., Deros, S.N.M., & Din, N.M. (2020). A review of vegetation encroachment detection in power transmission lines using optical sensing satellite imagery. *arXiv preprint arXiv:2010.01757*

He, C., Zhang, Y., Su, X., Xu, X., & Liao, M.-s. (2013). Target detection on high-resolution SAR image using Part-based CFAR Model. In, *2013 IEEE International Geoscience and Remote Sensing Symposium-IGARSS* (pp. 3570-3573): IEEE

He, K., Zhang, X., Ren, S., & Sun, J. (2016). Deep residual learning for image recognition. In, *Proceedings of the IEEE conference on computer vision and pattern recognition* (pp. 770-778)

He, W., Chen, W., Chen, B., Yang, S., Xie, D., Lin, L., Qi, D., & Zhuang, Y. (2023). Unsupervised Prompt Tuning for Text-Driven Object Detection. In, *Proceedings of the IEEE/CVF International Conference on Computer Vision* (pp. 2651-2661)

Hou, L., Lu, K., Xue, J., & Li, Y. (2022). Shape-adaptive selection and measurement for oriented object detection. In, *Proceedings of the AAAI Conference on Artificial Intelligence* (pp. 923-932)

Khattak, M.U., Rasheed, H., Maaz, M., Khan, S., & Khan, F.S. (2023). Maple: Multi-modal prompt learning. In, *Proceedings of the IEEE/CVF Conference on Computer Vision and Pattern Recognition* (pp. 19113-19122)

Kirillov, A., Mintun, E., Ravi, N., Mao, H., Rolland, C., Gustafson, L., Xiao, T., Whitehead, S., Berg, A.C., & Lo, W.-Y. (2023). Segment anything. *arXiv preprint arXiv:2304.02643*

Li, J., Li, Y., Jiang, H., & Zhao, Q. (2022a). Hierarchical Transmission Tower Detection from High-Resolution SAR Image. *Remote Sensing, 14*, 625

Li, T., Wang, C., Zhu, S., Zou, L., Wu, F., Xu, L., & Zhang, H. (2023). Power Transmission Tower Detection Based on High-Resolution SAR Images and Deep Learning. In, *2023 SAR in Big Data Era (BIGSARDATA)* (pp. 1-4): IEEE

Li, W., Chen, Y., Hu, K., & Zhu, J. (2022b). Oriented reppoints for aerial object detection. In, *Proceedings of the IEEE/CVF conference on computer vision and pattern recognition* (pp. 1829-1838)

Lin, T.-Y., Goyal, P., Girshick, R., He, K., & Dollár, P. (2017). Focal loss for dense object detection. In, *Proceedings of the IEEE international conference on computer vision* (pp. 2980-2988)

Liu, L., Du, R., & Liu, W. (2019). Flood distance algorithms and fault hidden danger recognition for transmission line towers based on SAR images. *Remote Sensing, 11*, 1642

Liu, Y., Wu, Z.-X., Xu, L.-G., & Zhang, W.-F. (2012). High voltage transmission-tower detection method based on high-resolution remote sensing image. *Journal of Electric Power, Science, and Technology, 27*, 47-51

Liu, Z., Lin, Y., Cao, Y., Hu, H., Wei, Y., Zhang, Z., Lin, S., & Guo, B. (2021). Swin transformer: Hierarchical vision transformer using shifted windows. In, *Proceedings of the IEEE/CVF international conference on computer vision* (pp. 10012-10022)

Long, Y., Han, J., Huang, R., Xu, H., Zhu, Y., Xu, C., & Liang, X. (2023). Fine-Grained Visual–Text Prompt-Driven Self-Training for Open-Vocabulary Object Detection. *IEEE Transactions on Neural Networks and Learning Systems*

Lu, Y., Liu, J., Zhang, Y., Liu, Y., & Tian, X. (2022). Prompt distribution learning. In, *Proceedings of the IEEE/CVF Conference on Computer Vision and Pattern Recognition* (pp. 5206-5215)

Ma, J., Shao, W., Ye, H., Wang, L., Wang, H., Zheng, Y., & Xue, X. (2018). Arbitrary-oriented scene text detection via rotation proposals. *IEEE transactions on multimedia, 20*, 3111-3122

Matikainen, L., Lehtomäki, M., Ahokas, E., Hyyppä, J., Karjalainen, M., Jaakkola, A., Kukko, A., & Heinonen, T. (2016). Remote sensing methods for power line corridor surveys. *ISPRS Journal of Photogrammetry and Remote Sensing, 119*, 10-31

Ming, Q., Miao, L., Zhou, Z., Yang, X., & Dong, Y. (2021). Optimization for arbitrary-oriented object detection via representation invariance loss. *IEEE Geoscience and Remote Sensing Letters, 19*, 1-5

Muhammad, M.B., & Yeasin, M. (2020). Eigen-cam: Class activation map using principal components. In, *2020 international joint conference on neural networks (IJCNN)* (pp. 1-7): IEEE

Nie, X., Ni, B., Chang, J., Meng, G., Huo, C., Xiang, S., & Tian, Q. (2023). Pro-tuning: Unified prompt tuning for vision tasks. *IEEE Transactions on Circuits and Systems for Video Technology*

Redmon, J., Divvala, S., Girshick, R., & Farhadi, A. (2016). You only look once: Unified, real-time object detection. In, *Proceedings of the IEEE conference on computer vision and pattern recognition* (pp. 779-788)

Ren, S., He, K., Girshick, R., & Sun, J. (2015). Faster r-cnn: Towards real-time object detection with region proposal networks. *Advances in neural information processing systems, 28*

Tancik, M., Srinivasan, P., Mildenhall, B., Fridovich-Keil, S., Raghavan, N., Singhal, U., Ramamoorthi, R., Barron, J., & Ng, R. (2020). Fourier features let networks learn high frequency functions in low dimensional domains. *Advances in neural information processing systems, 33*, 7537-7547

Tian, G., Meng, S., Bai, X., Zhi, Y., Ou, W., Fei, X., & Tan, Y. (2020). Electric tower target identification based on high-resolution SAR image and deep learning. In, *Journal of Physics: Conference Series* (p. 012117): IOP Publishing

Touvron, H., Cord, M., Douze, M., Massa, F., Sablayrolles, A., & Jégou, H. (2021). Training data-efficient image transformers & distillation through attention. In, *International conference on machine learning* (pp. 10347-10357): PMLR

Union, G.I.O. (2019). A metric and a loss for bounding box regression. In, *Rezatofighi, N. Tsoi, J. Gwak, A. Sadeghian, I. Reid, S. Savarese//IEEE/CVF Conference on Computer Vision and Pattern Recognition (CVPR), Long Beach, CA, USA* (pp. 658-666)

Wang, K., Li, C., Tu, Z., & Luo, B. (2023). Unified-modal Salient Object Detection via Adaptive Prompt Learning. *arXiv preprint arXiv:2311.16835*

Wang, W., Xie, E., Li, X., Fan, D.-P., Song, K., Liang, D., Lu, T., Luo, P., & Shao, L. (2021). Pyramid vision transformer: A versatile backbone for dense prediction without convolutions. In, *Proceedings of the IEEE/CVF international conference on computer vision* (pp. 568-578)

Wang, W., Xie, E., Li, X., Fan, D.-P., Song, K., Liang, D., Lu, T., Luo, P., & Shao, L. (2022). Pvt v2:

Improved baselines with pyramid vision transformer. *Computational Visual Media, 8*, 415-424

Wu, B., Wang, H., & Chen, J. (2023). Feature Enhancement Using Multi-Baseline SAR Interferometry-Correlated Synthesis Images for Power Transmission Tower Detection in Mountain Layover Area. *Remote Sensing, 15*, 3823

Xie, E., Wang, W., Yu, Z., Anandkumar, A., Alvarez, J.M., & Luo, P. (2021a). SegFormer: Simple and efficient design for semantic segmentation with transformers. *Advances in neural information processing systems, 34*, 12077-12090

Xie, X., Cheng, G., Wang, J., Yao, X., & Han, J. (2021b). Oriented R-CNN for object detection. In, *Proceedings of the IEEE/CVF international conference on computer vision* (pp. 3520-3529)

Yan, L., Wu, W., & Li, T. (2011). Power transmission tower monitoring technology based on TerraSAR-X products. In, *International Symposium on Lidar and Radar Mapping 2011: Technologies and Applications* (pp. 374-380): SPIE

Yang, X., & Yan, J. (2020). Arbitrary-oriented object detection with circular smooth label. In, *Computer Vision–ECCV 2020: 16th European Conference, Glasgow, UK, August 23–28, 2020, Proceedings, Part VIII 16* (pp. 677-694): Springer

Yang, X., Yan, J., Feng, Z., & He, T. (2021a). R3det: Refined single-stage detector with feature refinement for rotating object. In, *Proceedings of the AAAI conference on artificial intelligence* (pp. 3163-3171)

Yang, X., Yan, J., Ming, Q., Wang, W., Zhang, X., & Tian, Q. (2021b). Rethinking rotated object detection with gaussian wasserstein distance loss. In, *International conference on machine learning* (pp. 11830-11841): PMLR

Yang, X., Yang, J., Yan, J., Zhang, Y., Zhang, T., Guo, Z., Sun, X., & Fu, K. (2019). Scrdet: Towards more robust detection for small, cluttered and rotated objects. In, *Proceedings of the IEEE/CVF international conference on computer vision* (pp. 8232-8241)

Yang, X., Yang, X., Yang, J., Ming, Q., Wang, W., Tian, Q., & Yan, J. (2021c). Learning high-precision bounding box for rotated object detection via kullback-leibler divergence. *Advances in neural information processing systems, 34*, 18381-18394

Zeng, T., Gao, Q., Ding, Z., Tian, W., Yang, Y., & Zhang, Z. (2017). Power transmission tower detection based on polar coordinate semivariogram in high-resolution SAR image. *IEEE Geoscience and Remote Sensing Letters, 14*, 2200-2204

Zha, W., Hu, L., Duan, C., & Li, Y. (2023). Semi-supervised learning-based satellite remote sensing object detection method for power transmission towers. *Energy Reports, 9*, 15-27

Zhang, H., Li, F., Liu, S., Zhang, L., Su, H., Zhu, J., Ni, L.M., & Shum, H.-Y. (2022). Dino: Detr with improved denoising anchor boxes for end-to-end object detection. *arXiv preprint arXiv:2203.03605*

Zhang, J., & Xie, Q. (2019). Failure analysis of transmission tower subjected to strong wind load. *Journal of Constructional Steel Research, 160*, 271-279

Zhang, P., Li, Z., & Chen, Q. (2013a). Detection of power transmission tower from SAR image based on the fusion method of CFAR and EF feature. In, *2013 IEEE International Geoscience and Remote Sensing Symposium-IGARSS* (pp. 4018-4021): IEEE

Zhang, S., Chi, C., Yao, Y., Lei, Z., & Li, S.Z. (2020). Bridging the gap between anchor-based and anchor-free detection via adaptive training sample selection. In, *Proceedings of the IEEE/CVF conference on computer vision and pattern recognition* (pp. 9759-9768)

Zhang, X., Zhou, Z., Gong, H., Lu, S., Tian, D., & Institute, W.N.L.L.C.o.S.G.E.P.R. (2013b). Power transmission-tower detection for SAR images based on SIRV model. *Eng. J. Wuhan Univ., 46*, 737-741

Zhou, K., Yang, J., Loy, C.C., & Liu, Z. (2022a). Learning to prompt for vision-language models. *International Journal of Computer Vision, 130*, 2337-2348

Zhou, Y., Sheng, Q., Chen, J., Li, N., Fu, X., & Zhou, Y. (2022b). The failure mode of transmission tower foundation on the landslide under heavy rainfall: a case study on a 500-kV transmission tower foundation on the Yanzi landslide in Badong, China. *Bulletin of Engineering Geology and the Environment, 81*, 125

Zhou, Y., Yang, X., Zhang, G., Wang, J., Liu, Y., Hou, L., Jiang, X., Liu, X., Yan, J., & Lyu, C. (2022c). Mmrotate: A rotated object detection benchmark using pytorch. In, *Proceedings of the 30th ACM International Conference on Multimedia* (pp. 7331-7334)

Zhu, X., Su, W., Lu, L., Li, B., Wang, X., & Dai, J. (2020). Deformable detr: Deformable transformers for end-to-end object detection. *arXiv preprint arXiv:2010.04159*